\documentclass[runningheads]{llncs}
\usepackage[T1]{fontenc}
\usepackage{times,latexsym}
\usepackage{url}

\usepackage{cite}
\usepackage{amsmath,amssymb,amsfonts}
\usepackage{algorithmic}
\usepackage{graphicx}
\usepackage{textcomp}
\usepackage{xcolor}
\usepackage{multirow}   
\usepackage{booktabs}   
\usepackage{array}      
\usepackage{amssymb} 
\usepackage[normalem]{ulem}
\usepackage[table]{xcolor} 
\usepackage{pifont}   
\usepackage{url}
\usepackage{subcaption}
\usepackage{hyperref}
\usepackage{fontawesome5} 

\usepackage{rotating}
\usepackage{multirow}
\usepackage{booktabs}
\usepackage{colortbl}
\usepackage{array}
\usepackage{pdflscape}   
\usepackage{rotating}     
\begin{document}
%


\title{\textsc{AIssistant}: Human-AI Collaborative Review and Perspective Research Workflows in Data Science}

%
%
%
\author{Sasi Kiran Gaddipati\inst{1}\orcidID{0000-0003-3098-4592}\thanks{Equal contribution.} \and
Farhana Keya \inst{1}\orcidID{0000-0002-3782-8069}\protect\footnotemark[1] \and
Gollam Rabby\inst{1, 2}\orcidID{0000-0002-1212-0101}\thanks{Corresponding author.} \and
Sören Auer \inst{1, 2}\orcidID{0000-0002-0698-2864}}
\authorrunning{S. K. Gaddipati et al.}
\institute{TIB Leibniz Information Centre for Science and Technology, 30167 Hanover, Germany \\
\email{\{sasi.gaddipati, farhana.keya, gollam.rabby, auer\}@tib.eu} \and
L3S Research Center, 30167 Hanover, Germany}

\maketitle   
\begin{abstract}

High-quality scientific review and perspective papers require substantial time and effort, limiting researchers' ability to synthesize emerging knowledge. While Large Language Models (LLMs) leverage AI Scientists for scientific workflows, existing frameworks focus primarily on autonomous workflows with very limited human intervention. We introduce \textsc{AIssistant}, the first open-source agentic framework for Human--AI collaborative generation of scientific perspectives and review research in data science. \textsc{AIssistant} employs specialized LLM-driven agents augmented with external scholarly tools and allows human intervention throughout the workflow. The framework consists of two main multi-agent systems: \textit{Research Workflow} with seven agents and a \textit{Paper Writing Workflow} with  eight agents. We conducted a comprehensive evaluation with both human expert reviewers and LLM-based assessment following NeurIPS standards. Our experiments show that  \texttt{OpenAI o1} achieves the highest quality scores on chain-of-thought prompting with augmented Literature Search tools. We also conducted a Human--AI interaction survey with results showing a 65.7\% time savings. We believe that our work establishes a baseline for Human--AI collaborative scientific workflow for review and perspective research in data science, demonstrating that agent-augmented pipelines substantially reduce effort while maintaining research integrity through strategic human oversight.



\end{abstract}
%
%
%

\section{Introduction}

\begin{quote}
\itshape
“We can only see a short distance ahead, but we can see plenty there that needs to be done.” — Alan Turing.
\vspace{-2mm}
\end{quote}

\noindent Turing’s observation remains relevant to this day; the space of open scientific questions is vast, yet researchers can pursue only a small subset of them due to resource limitations. These constraints often restrict the development of high-impact perspective and review research that shape emerging developments in science. Recent advances in Large Language Models (LLMs) demonstrate promise in scientific workflows, including idea generation, literature synthesis, and drafting scholarly texts~\cite{lu2024ai}~\cite{yamada2025ai},~\cite{schmidgall2025Agentlaboratoryusingllm},~\cite{keya2025sci}. However, existing systems largely emphasize fully autonomous workflows and provide limited human oversight or integration with scholarly infrastructures. To address this gap, we introduce \textsc{AIssistant}\footnote{Github Link: \url{https://github.com/gsasikiran/TIB-webapp/}}, an open-source agentic framework that enables efficient human–AI collaborative perspectives and review research workflows in data science. Unlike other approaches, \textsc{AIssistant} employs specialized LLM-driven agents augmented with external tools and structured human feedback throughout the workflow. The main contributions of this paper are as follows:


\begin{table*}[t]
\centering
\caption{Comparison of recent AI Scientist systems on Human–AI collaboration for review and perspective research workflows. Rev. = Review paper support, Pers. = Perspective paper support, HITL = Human-in-the-Loop.}
\scriptsize
\setlength{\tabcolsep}{3pt}
\renewcommand{\arraystretch}{1.0}
\begin{tabular}{p{2.7cm} p{0.5cm} p{0.5cm} p{3.5cm} p{2.7cm} p{0.5cm}}
\toprule
\textbf{Name} & \textbf{Rev.} & \textbf{Pers.} & \textbf{Contribution} & \textbf{Approach} & \textbf{Year} \\ 
\midrule
AI Scientist-V1 \cite{lu2024ai} & \ding{55} & \ding{55} & Foundational frameworks & Symbolic reasoning & 2024 \\
AI Scientist-V2 \cite{yamada2025ai} & \ding{55} & \ding{55} & Interactive AI systems & Multi-Agent LLMs & 2025 \\
Agent Laboratory \cite{schmidgall2025Agentlaboratoryusingllm} & \ding{55} & \ding{55} & Multi-agent coordination & Agents + HITL & 2025 \\
Cycle Researcher \cite{weng2024cycleresearcher} & \ding{55} & \ding{55} & Paper generation LLM & Iterative RL & 2024 \\
Zochi \cite{zochi2025} & \ding{55} & \ding{55} & Bias/fairness focus & Ethical auditing & 2025 \\
Tiny Scientist \cite{yu2025tinyscientist} & \ding{55} & \ding{55} & Lightweight research agent & Multi-agent research & 2025 \\
\textbf{\textsc{AIssistant}} & \checkmark & \checkmark & Review/perspective workflow & Multi-agent + HITL & 2025 \\
\bottomrule
\end{tabular}
\label{tab:ai_collaboration}
\vspace{-0.8 cm}
\end{table*}

\begin{enumerate}
    \item We present \textsc{AIssistant}, the first open-source agentic framework for accelerating scientific perspective and review research workflows in data science through human-AI collaboration, integrating specialised LLM agents and external Literature Search tools (\texttt{LS tools}).
    \item We conduct human and automated evaluations across multiple LLMs, finding that \texttt{OpenAI o1} achieves the highest NeurIPS-style review-quality research paper generation, with strong alignment between LLM-based and expert human judgments.
    \item We provide detailed cost and inference-time analysis, demonstrating that \textsc{AIssistant} generates scientific papers for under \$1 (with a max of \$0.64 per perspective and \$0.90 per review using \texttt{o1}; \$0.0022 and \$0.0019 for perspective and review papers, respectively, using \texttt{gpt-4o-mini}).
\end{enumerate}

\noindent To our knowledge, this is the first systematic study of human–AI agentic frameworks for perspective and review research in data science. Results indicate that agent-augmented pipelines let researchers focus on creativity and reasoning while delegating structured workflow to LLM-driven agents. Differences with existing AI scientist frameworks are summarized in Table~\ref{tab:ai_collaboration}.

\section{Background \& Related Work}

Various LLMs, such as LLaMA~\cite{touvron2023llama}, Claude~\cite{TheC3}, etc, trained on massive datasets using transformer architectures~\cite{vaswani2017attention}, have demonstrated remarkable text generation capabilities. Advances in chain-of-thought reasoning have further improved both task performance and interpretability~\cite{jaech2024openai}, motivating growing interest in the extent to which AI can emulate human-level reasoning. However, standalone LLMs face challenges in solving complex, real-world problems, particularly in the scientific domain. Agentic frameworks address this limitation by coordinating multiple LLM agents that collaborate on individual tasks toward shared goals. Such frameworks have achieved promising results in domains ranging from medical diagnosis~\cite{mcduff2025towards}~\cite{schmidgall2024Agentclinic} and robotics~\cite{black2024pi_0} to software engineering~\cite{jimenez2023swe}~\cite{wang2025openhandsopenplatformai} and web automation~\cite{deng2023mind2web}. Collectively, these studies suggest that multi-agent systems have the potential to tackle problems beyond the capacity of a single LLM. The success of the agentic frameworks has inspired their integration into scientific discovery. Later studies have explored such integration in mathematics~\cite{romera2024mathematical}, chemistry
~\cite{m2024augmenting}, material science~\cite{merchant2023scaling}~\cite{zhang2024honeycomb}, biodiversity~\cite{ziegler2025ai}, and algorithm design~\cite{fawzi2022discovering}. While some studies suggest that LLMs can generate novel insights~\cite{si2024can}, others emphasize persistent limitations in originality and reasoning~\cite{zhou2024shared}, highlighting the need for systematic evaluation and benchmarks in scientific contexts.  More recently, multi-agent systems such as Agent Laboratory~\cite{schmidgall2025Agentlaboratoryusingllm}, AI Scientist~\cite{lu2024ai}~\cite{yamada2025ai}, Zochi~\cite{zochi2025}, and Tiny Scientist~\cite{yu2025tinyscientist} have targeted research-related tasks in data science. These systems demonstrate progress in ideation, literature synthesis, and draft generation, but evaluations remain limited: most rely on hand-picked examples, lack standardized benchmarks, and emphasize full automation without systematic human intervention.

\begin{figure*}[t!]
    \centering
    \includegraphics[
        width=0.85\textwidth,
    ]{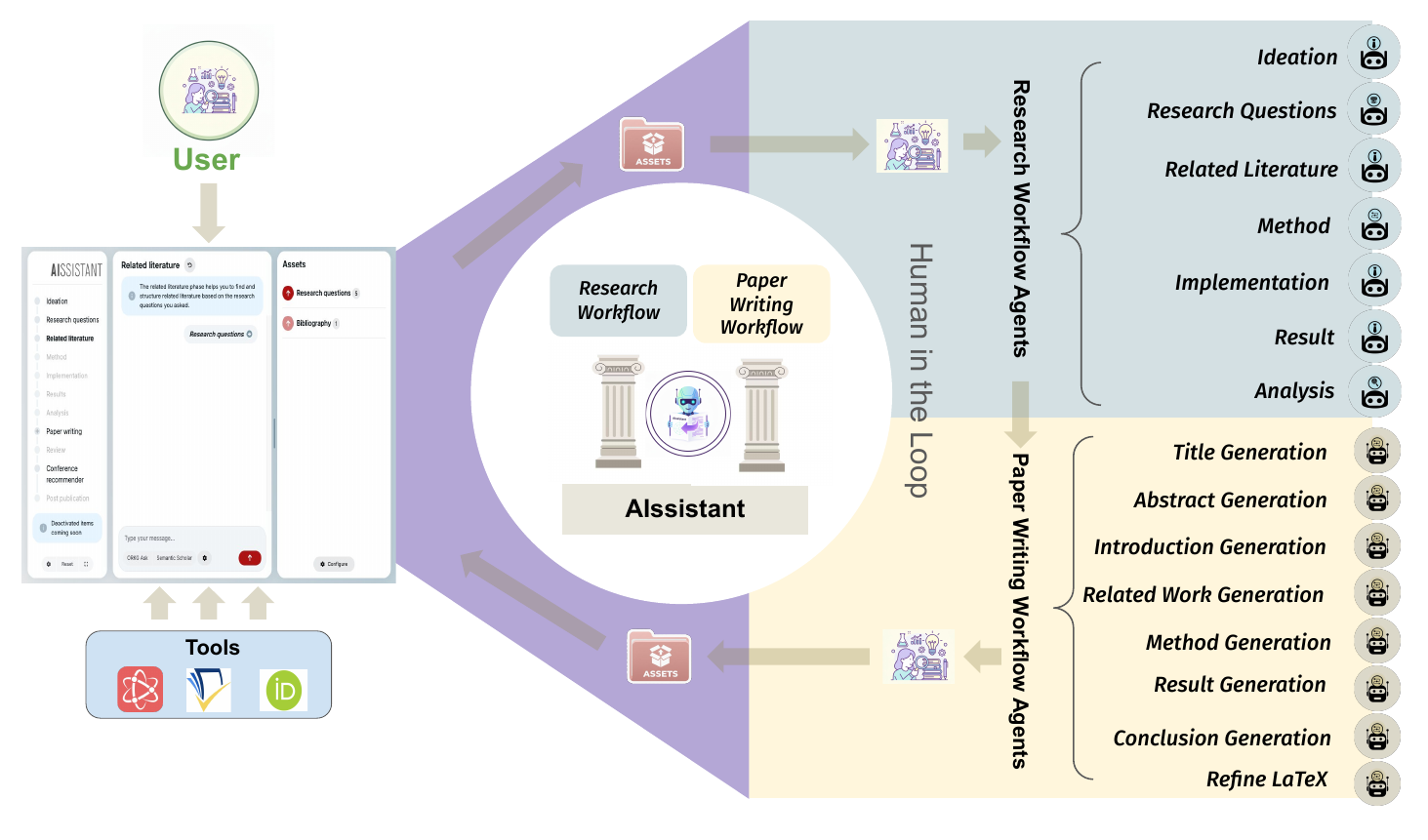}
    \caption{A schematic overview of the \textsc{AIssistant} framework, illustrating integrated research and paper-writing workflows with Human-in-the-Loop and flexible tool selection at each agent state.}
    
    \vspace{-0.5cm}
    \label{fig:Agentic_workflow}
\end{figure*}

\section{\textsc{AIssistant}}
\vspace{-0.25cm}

Our framework adopts a Human-in-the-Loop design to ensure accuracy, methodological rigor, adherence to scholarly standards, and responsibility to the user. The workflow is executed through a series of specialized agents. The \textsc{Research Workflow} comprises seven agents: \emph{Ideation}, \emph{Research Questions}, \emph{Related Literature}, \emph{Method}, \emph{Implementation}, \emph{Result}, and \emph{ Analysis}. The \textsc{Paper Writing Workflow} contains eight agents with tailored prompts for review and perspective papers: \emph{Title generation}, \emph{Abstract generation}, \emph{Introduction generation}, \emph{Related Work Generation}, \emph{Method \& Implementation Generation}, \emph{Result \& Discussion Generation}, \emph{Conclusion Generation}, and \emph{Refine \LaTeX{}} to finalize manuscripts into publication-ready form by improving coherence over all the individual agent-generated texts.  This design supports flexible agent selection while maintaining a coherent, traceable workflow. As illustrated in~\autoref{fig:Agentic_workflow}, human inputs, AI-generated texts, and intermediate human-selected outputs, which we refer to as \texttt{Assets}, propagate through the latter agents to produce comprehensive review and perspective research workflow generation.

\noindent \textbf{Task Formulation.}
Although each specialized agent in \textsc{AIssistant} performs a distinct workflow function, they follow a unified formal structure. We formulate the task function of an agent $i$, user input $U_i$, system prompt $P_i$, agent  tool set $T_i = \{t_{i1}, \dots, t_{in}\}$, generated outputs $M_i$, and asset state $A_i$ as $\mathcal{F}_i : (U_i, P_i, T_i, A_i^-) \rightarrow (M_i, A_i),$ where $A_i^-$ denotes assets available before execution and $A_i$ the updated assets of the current agent $i$ after human selection of outputs. In \textsc{Paper Writing Workflow}, all paper-generation assets are aggregated and revised by the \emph{Refine \LaTeX{}} agent to improve coherence, flow, and formatting. The bibliography is excluded to avoid hallucinated modifications of cited titles. Let $\{A_{i_1}, \dots, A_{i_k}\}$ be the relevant assets; the final manuscript processed through this multi-agent pipeline is $M_{\text{Final}} = 
\mathcal{F}_{\text{LLM}}\!\left(\bigcup_{j=1}^k A_{i_j} \setminus \text{Bibliography}\right),$ yielding a coherent, publication-ready output while preserving citation integrity, with a flexibility for further human modifications.

\noindent \textbf{Multi-Agent Pipeline.}
The multi-agent pipeline extends the task formulation $\mathcal{F}_i : (U_i, P_i, T_i, A_i^-) \to (M_i, A_i)$ by composing agents as $\mathcal{F}_{\text{pipeline}} = \mathcal{F}_{1} \circ \dots \circ \mathcal{F}_{N}$, with asset updates $A_i = \mathcal{H}(M_i \cup A_i^-)$ under human selection. The \emph{Research workflow} consists of seven agents $\{s_i,\dots,s_K\}$ for ideation, research questions, related literature, method, implementation, results, and analysis as shown in \autoref{fig:Agentic_workflow}; each agent follows $(M_{s_j}, A_{s_j}) = \mathcal{F}_{s_j}(U_{s_j}, P_{s_j}, T_{s_j}, A_{s_j}^-)$, producing aggregated assets $A_{\text{MD}} = \bigcup_{j=1}^{K} A_{s_j}$. The \emph{Paper Writing Workflow} applies a eight-step sequential $\mathcal{F}_{\text{RP}} = \mathcal{F}_{\text{TitleGen}} \circ \mathcal{F}_{\text{AbstractGen}} \circ \dots \circ \mathcal{F}_{\text{RefineLatex}}$ to generate a structured draft, refine coherence and formatting, and integrate human corrections, yielding the final manuscript, that can be editable by a human ($M_{\text{Final}} = \mathcal{H}(M_{\text{Edit}})$). Overall, the pipeline produces a publication-ready output through a mathematically grounded, human-in-the-loop multi-agent workflow: $M_{\text{final}} = \mathcal{F}_{\text{pipeline}}(U, P, T, A)$.

\section{Experimentation}

We generated a total of 48 papers (24 perspective papers, 24 review papers) using \textsc{AIssistant} with human-AI collaboration, seeding the \emph{Ideation Agent} with published paper titles to minimise domain drift and align with the knowledge cutoffs of the LLMs. Two PhD-level data science researchers executed the full \textsc{AIssistant} workflow, selecting outputs and propagating \texttt{Assets} across agents. All the agents are set homogeneously to either \texttt{OpenAI o1} or \texttt{gpt-4o-mini},  and the same prompt strategy, except for the \emph{Refine \LaTeX{}} agent, for which we use \texttt{OpenAI o1} with \textit{chain-of-thought} prompt always to ensure complex reasoning and cross-section consistency checks.  Each paper was produced through the two defined \textsc{Research Workflow} and \textsc{Paper Writing Workflow} agents, subsequently further refined by the researchers to ensure coherence, format, and citation accuracy. 

\noindent \emph{Related Work} agent in \textsc{Research Workflow}  utilizes tools such as Semantic Scholar\footnote{\url{https://www.semanticscholar.org/}} and ORKG ASK\footnote{\url{https://ask.orkg.org/}} for retrieving relevant publications. This agent retrieves abstracts and bibliographical information from the top relevant papers and provides the user with access to select the relevant literature to the \emph{Assets}.  We experiment with both enabling and disabling these \texttt{Literature Search (LS) tools}, in which case, we directly utilize the knowledge of LLMs to generate the relevant literature as checkboxes. This helps us understand the efficiency of using \texttt{LS tools} compared to the actual knowledge of LLMs, and we present these comparative results in Section \ref{sec:evaluation_and_analysis}.

\noindent \textbf{Hardware Configurations.} We evaluated the \textsc{AIssistant} performance on two system configurations: (i) 32 GB RAM, 512 GB storage, Intel Core i7-165H (22 CPUs); (ii) 16 GB RAM, 512 GB storage, Intel Core i5-1145G7 (8 cores), Intel Xe GPU. Both configurations executed \texttt{OpenAI o1} and \texttt{GPT-4o-mini}, which do not require GPU acceleration. No substantial performance differences were observed across setups.

\begin{table}[t]
\centering
\setlength{\tabcolsep}{0.9pt} 
\renewcommand{\arraystretch}{1.0} 
\caption{Human-based review of \textsc{AIssistant } set to two LLMs (\texttt{GPT-4o-mini} and \texttt{OpenAI o1}) on review and perspective papers, with and without LS tools, across prompting strategies: Zero-Shot (ZS), Few-Shot (FS), and Chain-of-Thought (CoT).}
\resizebox{\textwidth}{!}{
\begin{tabular}{l|ccc|ccc|ccc|ccc|ccc|ccc|ccc|ccc}
\toprule
\multirow{4}{*}{\textbf{Human Based Review}} 
 & \multicolumn{12}{c|}{\textbf{\texttt{GPT-4o-mini}}} 
 & \multicolumn{12}{c}{\textbf{\texttt{OpenAI o1}}} \\
\cmidrule(lr){2-13} \cmidrule(lr){14-25}
 & \multicolumn{6}{c|}{\textbf{Review Papers}} 
 & \multicolumn{6}{c|}{\textbf{Perspective Papers}} 
 & \multicolumn{6}{c|}{\textbf{Review Papers}} 
 & \multicolumn{6}{c}{\textbf{Perspective Papers}} \\
\cmidrule(lr){2-7} \cmidrule(lr){8-13} \cmidrule(lr){14-19} \cmidrule(lr){20-25}
 & \multicolumn{3}{c}{Without LS Tools} & \multicolumn{3}{c|}{With LS Tools} 
 & \multicolumn{3}{c}{Without LS Tools} & \multicolumn{3}{c|}{With LS Tools} 
 & \multicolumn{3}{c}{Without LS Tools} & \multicolumn{3}{c|}{With LS Tools} 
 & \multicolumn{3}{c}{Without LS Tools} & \multicolumn{3}{c}{With LS Tools} \\
\cmidrule(lr){2-4} \cmidrule(lr){5-7} \cmidrule(lr){8-10} \cmidrule(lr){11-13} 
\cmidrule(lr){14-16} \cmidrule(lr){17-19} \cmidrule(lr){20-22} \cmidrule(lr){23-25}
 & ZS & FS & CoT & ZS & FS & CoT 
 & ZS & FS & CoT & ZS & FS & CoT 
 & ZS & FS & CoT & ZS & FS & CoT 
 & ZS & FS & CoT & ZS & FS & CoT \\
\midrule
Paper format (4) $\uparrow$ & 1.70 & 1.70 & 1.80 & 1.50 & 1.70 & 1.70 & 1.60 & \textbf{2.50} & 2.00 & 2.25 & 2.40 & 2.00 & 1.50 & 1.20 & 1.20 & 2.30 & 1.80 & 1.70 & 1.80 & 2.00 & 2.25 & 2.00 & 2.25 & 2.40 \\
Citations relevancy (4) $\uparrow$ & 1.50 & 1.20 & 1.70 & 1.50 & 1.70 & 1.70 & 2.00 & 2.00 & 1.40 & 2.00 & 1.80 & 2.20 & 1.30 & 1.20 & 1.20 & 1.70 & 1.70 & 1.80 & 1.80 & 1.75 & 2.00 & \textbf{2.25} & \textbf{2.25} & 2.00 \\
Hallucinations $\downarrow$ & 0.50 & 0.70 & 0.70 & 0.50 & 0.70 & 0.70 & 1.00 & 0.75 & 0.80 & 0.80 & 0.80 & 0.80 & 1.00 & 0.80 &  0.80 & 0.50 & 0.70 & 0.70 & 0.75 & 0.75 & 0.75 & 0.50 & 0.50 & 0.50 \\
Ethical concerns $\downarrow$ & 0.00 & 0.00 & 0.00 & 0.00 & 0.00 & 0.00 & 0.80 & 0.50 & 0.75 & 0.50 & 0.75 & 0.66 & 0.00 & 0.00 & 0.00 & 0.00 & 0.00 & 0.00 & 0.75 & 0.75 & 0.75 & 0.50 & 0.50 & 0.50 \\
\bottomrule
\end{tabular}}

\label{tab:paper_formate}
\vspace{-0.4cm}
\end{table}

\begin{figure}[t]
    \centering
    \includegraphics[width=0.7\linewidth]{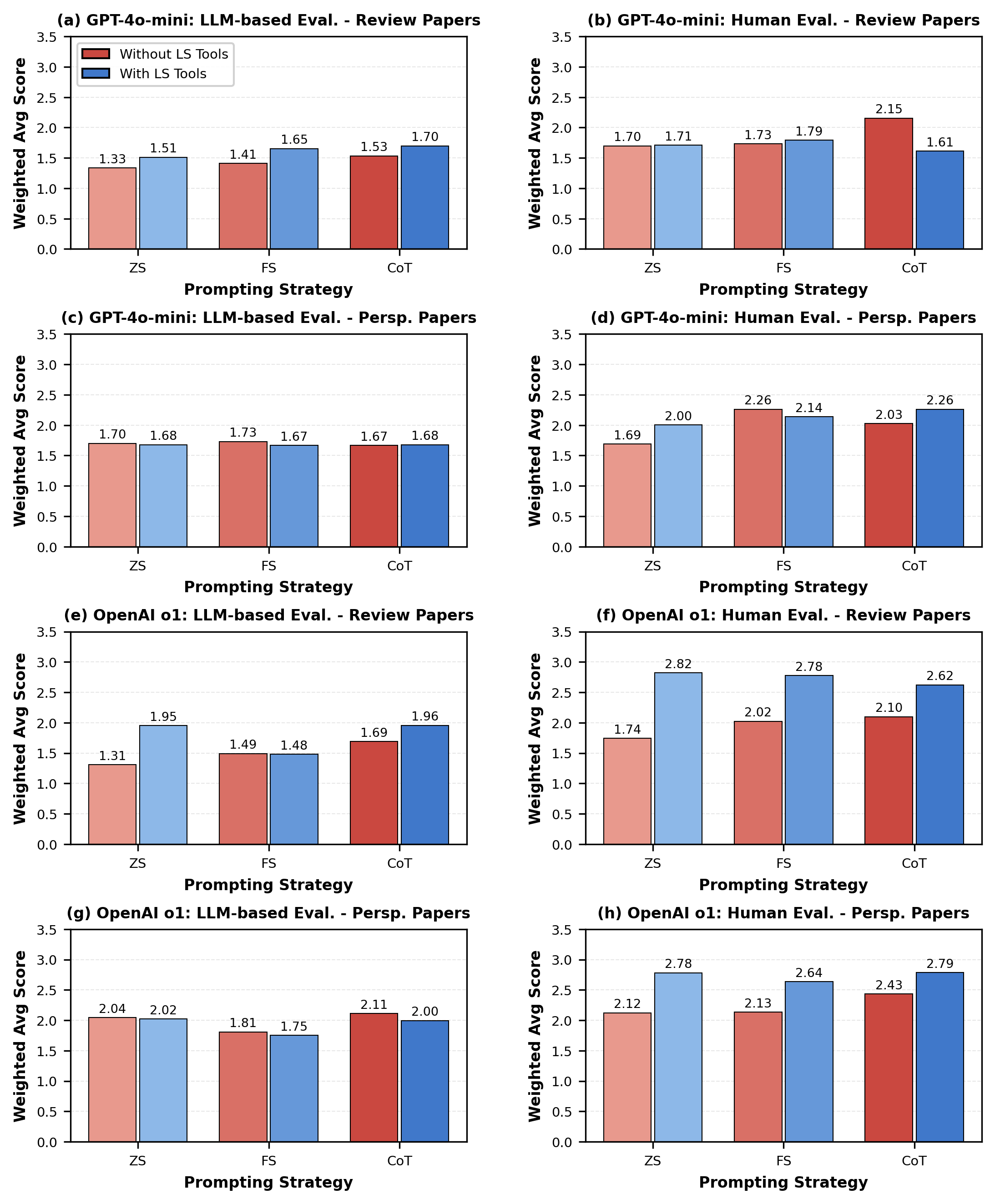}
    \caption{The weighted average scores of human and LLM-based evaluation of generated perspective and review papers using \textsc{AIssistant}. ZS = Zero Shot, FS = Few Shot, CoT = Chain-of-Thought, Eval. = Evaluation, Persp. = Perspective.}
    \vspace{-0.80cm}
    \label{fig:LLM_VS_Human}
\end{figure}

\begin{table*}[t]
\centering
\scriptsize
\setlength{\tabcolsep}{2pt} 
\renewcommand{\arraystretch}{1.0}

\caption{Program Chair insights highlighting the LLM and human evaluations across multiple criteria. Scores are shown as `Average (Range)'.}
\resizebox{\textwidth}{!}{

\begin{tabular}{p{1.6cm}|p{1.9cm}|p{2.0cm}|p{2.0cm}|p{2.2cm}|p{4.0cm}}
\hline
\textbf{Criterion} & \textbf{Eval.} & \textbf{Review Paper} & \textbf{Perspective Paper} & \textbf{Overall Trend} & \textbf{Program Chair Insight} \\ \hline

Soundness (4) & LLM vs. Human & 1.50 (1.00-3.00) & 1.68 (1.00-3.00) & Slightly better with LLM & LLMs catch flaws; humans spot logical consistency \\ \hline
Presentation (4) & LLM vs. Human & 1.69 (1.00-3.00) & 2.07 (1.00-3.00) & More consistent with humans & LLMs excel in clarity, useful for initial reviews \\ \hline
Quality (4) & LLM vs. Human & 1.62 (1.00-3.00) & 1.80 (1.00-3.00) & Slightly better for humans & Humans excel at creativity and novelty \\ \hline
Clarity (4) & LLM vs. Human & 2.01 (1.00-3.00) & 2.20 (1.00-3.00) & Balanced across evaluations & Both LLMs and humans align in review writing quality \\ \hline
Significance (4) & LLM vs. Human & 1.74 (1.00-3.00) & 2.07 (1.00-3.00) & Slightly higher for humans & Humans have an edge in assessing impact and contribution \\ \hline
Originality (4) & LLM vs. Human & 1.69 (1.00-3.00) & 1.98 (1.00-3.00) & Similar ratings across methods & Humans excel in spotting transformative ideas \\ \hline
Overall (6) & LLM vs. Human & 2.22 (1.00-3.00) & 2.31 (1.00-3.50) & Slightly higher for humans & Humans provide deeper insights, especially for groundbreaking work \\ \hline

\end{tabular}}
\label{tab:Program-chair-analysis-summary}
\end{table*}

\begin{figure}[t]
    \centering
    \includegraphics[width=0.9\linewidth]{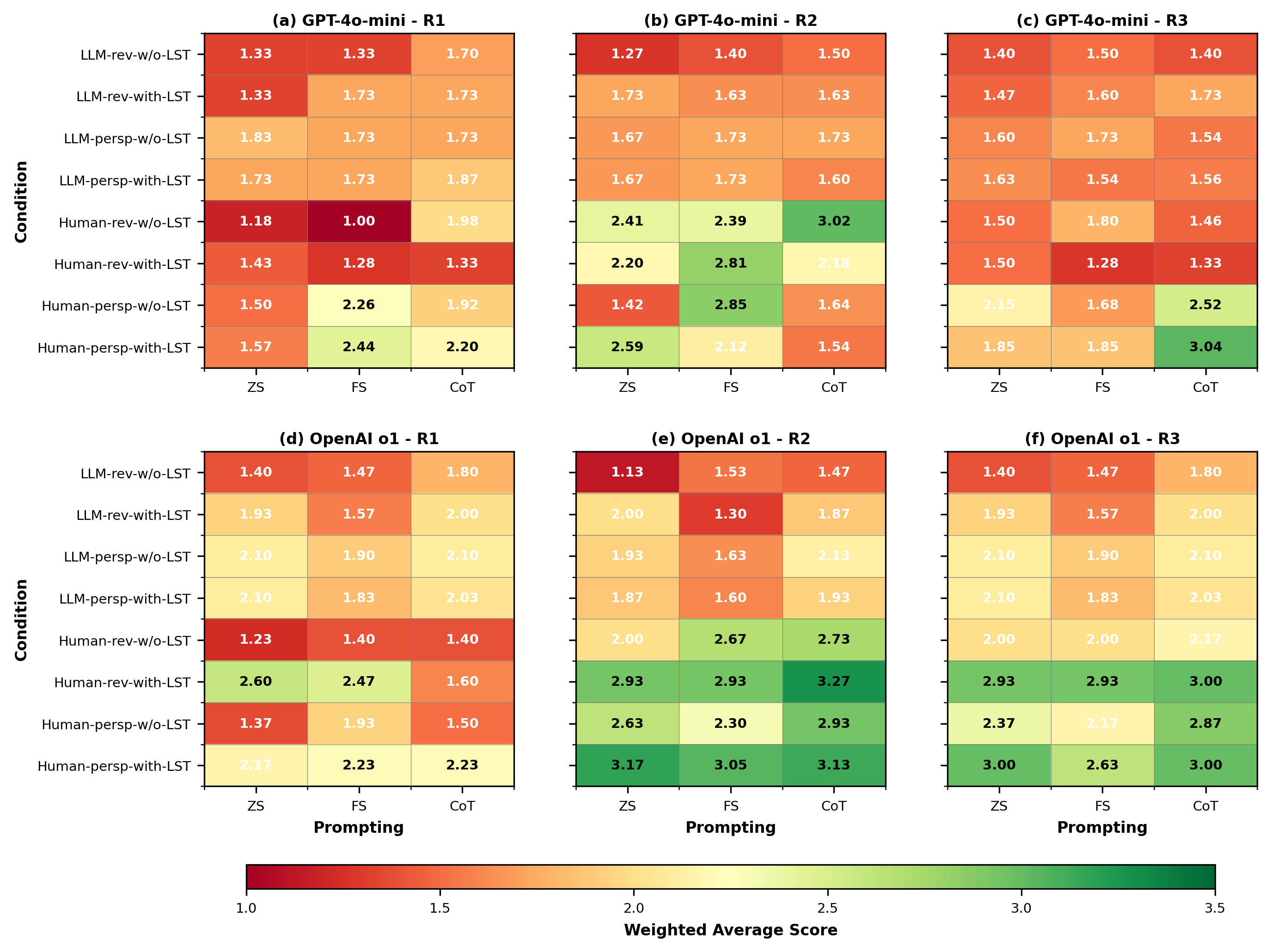}
    \vspace{-0.3cm}
    \caption{Human and LLM-based evaluation score distributions across all conditions. The y-axis shows combination of the each condition: LLM = LLM-based evaluation, Human = Human evaluation, rev = Review papers, persp = Perspective papers, LST = Literature Search Tools, R1 = Reviewer 1, R2 = Reviewer 2, R3 = Reviewer 3.}
    \label{fig:reviewer_heatmap}
    \vspace{-0.4cm}
\end{figure}

\begin{figure*}[t]
    \centering
    \includegraphics[width=0.8\textwidth]{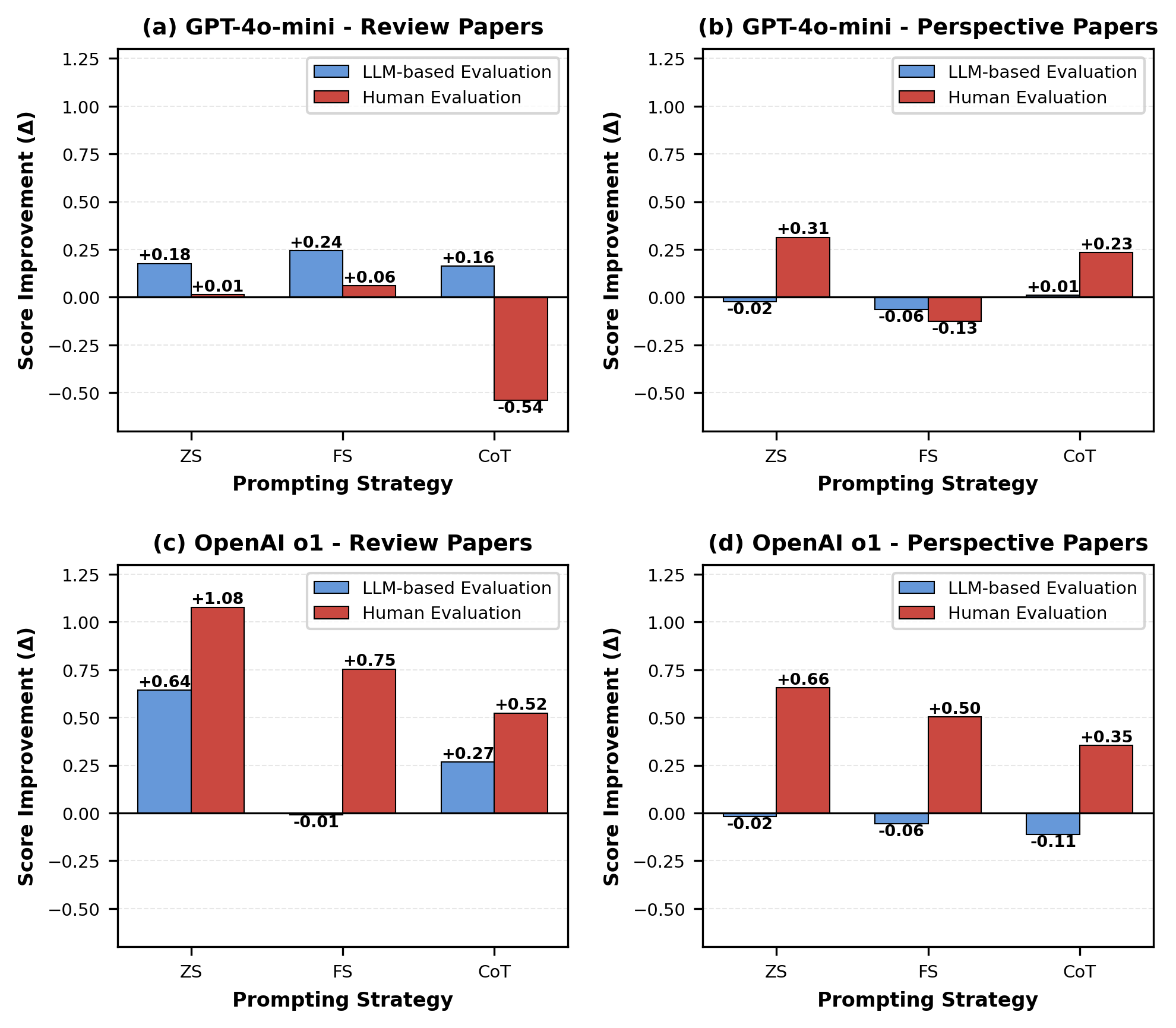}
    \caption{Comparison of score Improvement ($\Delta$) by LLM-based evaluation and human evaluation across prompting strategies. Panels (a) and (b) show results for the \texttt{GPT-4o-mini}, while (c) and (d) show results for the \texttt{OpenAI~o1}, across both Review and Perspective paper types. Human evaluation generally shows positive improvement, particularly for the \texttt{OpenAI~o1}, demonstrating the value of human collaboration in the proposed \textsc{AIssistant} framework.}
    \label{fig:improvement_comparison}
    \vspace{-0.7cm}
\end{figure*}

\section{Evaluation and Analysis}
\label{sec:evaluation_and_analysis}

To assess the quality, reliability, and usability of the \textsc{AIssistant}, we conducted a comprehensive evaluation combining human expert reviews, LLM-based assessments 
and human-computer interaction studies
. We submitted 48 papers for evaluation, following a strict double-blind protocol. While the dataset size was constrained to ensure feasibility, each human reviewer assessed all submissions; it remains larger and more balanced than prior AI Scientist benchmarks~\cite{schmidgall2025Agentlaboratoryusingllm, yamada2025ai}.

\subsection{Human Reviewers Evaluation}
\label{sec:human_reviewers_evaluation}
To ensure rigor, 48 papers were evaluated by independent reviewers in accordance with NeurIPS standards. The authors did not participate to avoid bias. Reviewers were selected based on expertise, publication record, and conflict-of-interest checks. Each manuscript was scored on clarity, originality, technical soundness, significance, reproducibility, limitations, and ethical considerations; missing sections were penalized. Hallucinations, fabricated evidence, and ethically sensitive content were flagged. Each paper received three independent reviews; scores, qualitative feedback, and reviewer confidence (1--5) were aggregated, interpolating missing values for consistency. \autoref{tab:paper_formate} and \autoref{fig:LLM_VS_Human} summarize evaluations for \texttt{GPT-4o-mini} and \texttt{OpenAI~o1} under \emph{With LS Tools} and \emph{Without LS Tools} for ZS, FS, and CoT prompting. \texttt{OpenAI~o1} showcased the best human-rated results on both perspective and review paper research utilizing \texttt{LS tools}, achieving  weighted averages of 2.79 and 2.82, respectively. However, in the  case of perspective papers, the best  weighted average is achieved through CoT prompting, while in review paper research, the best score is achieved from zero-shot prompting. The best-rated human-reviewed paper with \texttt{GPT-4o-mini} generated research is 2.26, for perspective, compared to the 2.15 for review paper research. Overall, from human evaluation feedback, the chair observed that LS tools with CoT prompting improved \texttt{OpenAI~o1} across structure, citation relevance, and hallucination reduction.

\subsection{LLM Reviewers Evaluation}
We employed \texttt{GPT-5} ~\cite{openai2025gpt5} as an agent-based reviewer to obtain scalable and reproducible evaluations. GPT-5 followed the same rubric as human reviewers, as explained in Section \ref{sec:human_reviewers_evaluation},  generating three independent reviews per paper with prompts emphasizing feasibility, clarity, and novelty, respectively, mirroring the human workflow. For both review and perspective research, GPT-5 as a reviewer was critical compared to human reviewers. One reason could be the change in the distribution of the papers by (\texttt{OpenAI o1} and \texttt{gpt-4o-mini}) and the reviewing LLM was (\texttt{GPT-5}). The best results from LLM as a reviewer have been achieved with the perspective papers generated using \texttt{OpenAI o1} with CoT prompting and without using \texttt{LS tools}, yielding a weighted average of 2.11. This suggests that \texttt{GPT-5} as a reviewer cannot differentiate between real and hallucinated literature reviews. However, the LLM-based evaluation for the perspective papers has been better overall compared to the review papers, for which the best attained weighted average was 1.96 for the review research generated with \texttt{OpenAI o1} with CoT prompting. Human and LLM-based evaluation score distributions across all conditions are shown in \autoref{fig:reviewer_heatmap}.

\subsection{Program Chair Observation}
A program chair with 6-7 years of experience reviewing PhD-track and providing expert evaluation complementary to our LLM and human assessments. For each criterion, we report: (i) \texttt{Average} (mean score across LLMs and tool-supported variants), (ii) \texttt{Range} (difference between highest and lowest scores), and (iii) \texttt{Overall Trend} (LLM vs. human comparison), summarizing strengths, weaknesses, and alignment without listing individual scores. The chair observed consistent patterns: \texttt{Soundness} received the lowest ratings, averaging 1.50 for review and 1.68 for perspective papers, with LLMs slightly more positive than humans. The highest averages were for review and perspective papers overall (2.22 and 2.31), where humans scored higher. Other criteria—including presentation, quality, clarity, significance, and originality ranged 1.60–2.20. Humans consistently rated papers higher than \texttt{GPT-5}, summarized in~\autoref{tab:Program-chair-analysis-summary}. Citation relevance was similar across LLMs, with \texttt{OpenAI~o1} highest in FS and ZS. Hallucinations decreased notably for \texttt{OpenAI~o1} with \texttt{LS Tools}, especially in perspective papers. Ethical concerns were absent for review papers, while occasional hallucinated references were observed in perspective papers.

\subsection{Effectiveness of Human-AI Collaboration: Score Improvement}
\label{sec:score_improvement}

The Score Improvement ($\Delta$) analysis, as depicted in \autoref{fig:improvement_comparison}, reveals a complex and contingent relationship between review method, LLM choice, and prompting strategy. Across the four scenarios, human evaluation consistently achieves the maximum positive score improvement, peaking substantially for the \texttt{OpenAI~o1} on Review Papers with a ZS prompt ($\Delta=+1.08$), clearly demonstrating the superior potential of human-augmented feedback. However, human evaluation is also shown to be volatile, yielding the largest negative outcome ($\Delta=-0.54$) for the \texttt{GPT-4o-mini} using CoT, suggesting that human oversight without proper structuring can be detrimental. Overall, while both LLMs show competitive, positive improvements under certain conditions (e.g., \texttt{GPT-4o-mini} LLM-based evaluation on review papers), the results underscore that the greatest quality gains in the \textsc{AIssistant} framework are achieved through targeted human collaboration, specifically when paired with the stronger LLM (\texttt{OpenAI~o1}).

\begin{table*}[t]
\centering
\scriptsize
\caption{Human Evaluation metrics for \texttt{AIssistant} system acceptance, quality, and NASA Task Load Index(NASA-TLX). The ($\uparrow$) defines higher the better, and the ($\downarrow$) defines the score, the lower the better.}
\vspace{2mm}
\begin{minipage}{0.3\linewidth}
\centering
\textbf{System Acceptance (1--5)}
\begin{tabular}{lccc}
\toprule
Item & Mean & Med & SD \\
\midrule
Ease of use ($\uparrow$) & 4.0 & 4 & 0.63 \\
Complexity ($\downarrow$) & 1.8 & 2 & 0.75 \\
User confidence ($\uparrow$) & 3.2 & 3 & 0.98 \\
Future use intent ($\uparrow$) & 3.6 & 4 & 0.49 \\
\bottomrule
\end{tabular}
\end{minipage}
\hfill
\begin{minipage}{0.35\linewidth}
\centering
\textbf{NASA-TLX Workload Metrics (1--7)}
\begin{tabular}{lccc}
\toprule
Item & Mean & Med & SD \\
\midrule
Mental demand ($\downarrow$) & 4.0 & 4 & 1.41 \\
Effort ($\downarrow$) & 4.2 & 4 & 2.39 \\
Frustration ($\downarrow$) & 2.4 & 2 & 2.19 \\
Performance ($\uparrow$) & 5.8 & 6 & 0.45 \\
Temporal demand ($\downarrow$) & 3.2 & 3 & 0.75 \\
\bottomrule
\end{tabular}
\end{minipage}
\begin{minipage}{0.3\linewidth}
\centering
\textbf{System Quality Metrics (1--5)}
\begin{tabular}{lccc}
\toprule
Item & Mean & Med & SD \\
\midrule
Status visibility ($\uparrow$) & 3.2 & 3 & 0.84 \\
Real-world match ($\uparrow$) & 3.2 & 3 & 0.84 \\
User control ($\uparrow$) & 2.6 & 3 & 0.55 \\
Error prevention ($\uparrow$) & 2.4 & 2 & 1.14 \\
Minimalist design ($\uparrow$) & 3.8 & 4 & 0.45 \\
\bottomrule
\end{tabular}
\end{minipage}
\label{tab:UX-summary}
\vspace{-0.6cm}
\end{table*}

\subsection{Human-Computer Interaction (HCI) Analysis}

We conducted a Human--AI collaboration study with five expert users holding advanced degrees in data science and substantial experience with research workflows. While modest in size, this cohort aligns with expert-based evaluations commonly used in HCI and computer-supported cooperative work (CSCW), offering sufficient depth to surface usability patterns and interaction bottlenecks. Participants performed multiple end-to-end runs and provided detailed assessments of system behavior and Human--AI interaction. Acceptance metrics indicate strong usability: ease of use $\bar{x}=4.0$, complexity $\bar{x}=1.8$, and future use intent $\bar{x}=3.6$ (5-point scale). Notable efficiency gains were observed, with task duration reduced from 99 minutes (manual workflow) to 34 minutes with \textsc{AIssistant}, yielding a 65.7\% time savings. NASA--TLX measures further confirm usability, with mental demand $\bar{x}=4.0$, effort $\bar{x}=4.2$, frustration $\bar{x}=2.4$, and performance $\bar{x}=5.8$ (7-point scale). Overall, \textsc{AIssistant} demonstrates substantial efficiency improvements, low cognitive workload, and high user satisfaction. Consistent expert responses highlight its effectiveness for structured human--AI research workflows, with future work expanding to larger expert cohorts for broader validation. A detailed summary of user feedback is reported in \autoref{tab:UX-summary}.

\begin{figure*}[t]
    \centering
    \begin{subfigure}[t]{0.48\linewidth}
        \centering
        \includegraphics[width=\linewidth, height=0.2\textheight, keepaspectratio]{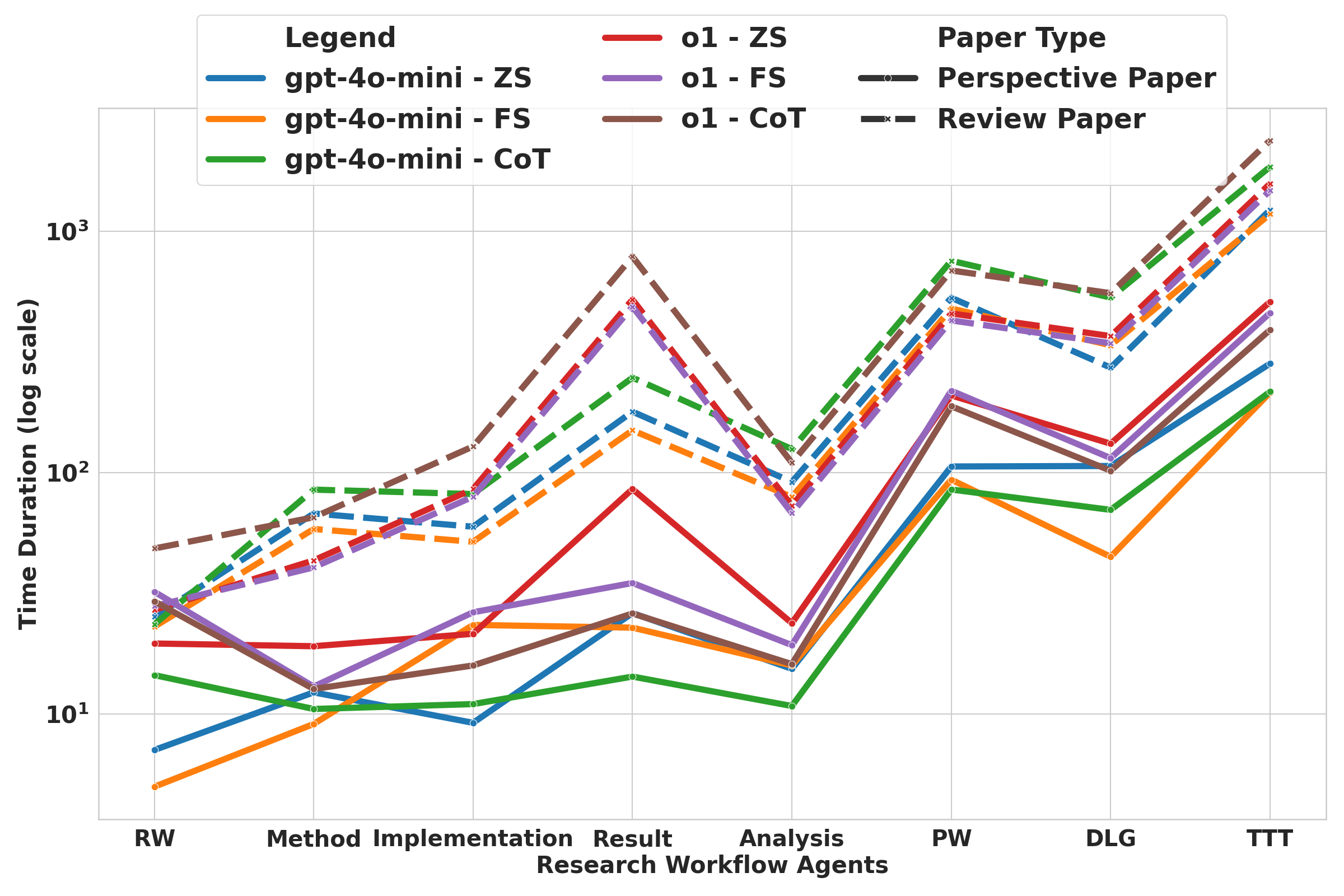}
        \caption{Inference time distribution for perspective and review papers, highlighting effort variability in a Human-AI workflow.}
        \label{fig:time_duration_barchart}
    \end{subfigure}%
    \hfill
    \begin{subfigure}[t]{0.48\linewidth}
        \centering
        \includegraphics[width=\linewidth, height=0.2\textheight, keepaspectratio]{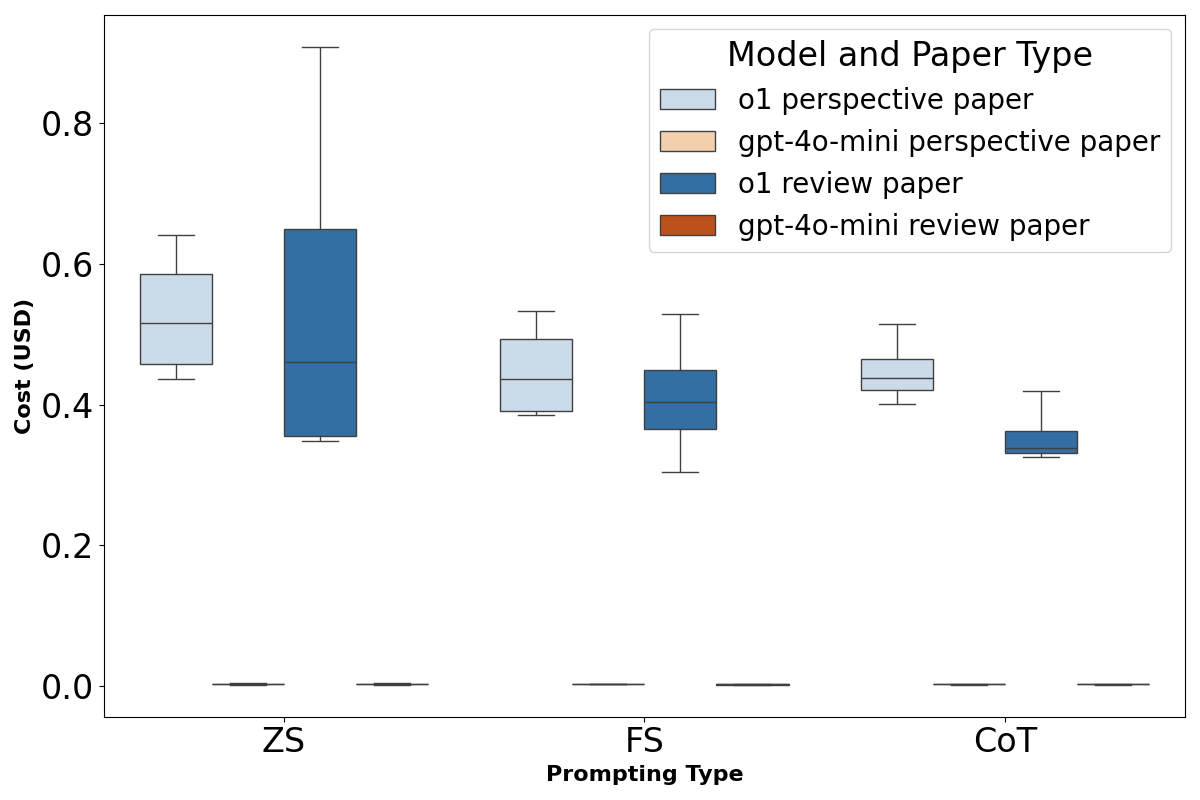}
        \caption{Comparison of costs (USD) across \texttt{GPT-4o-mini} and \texttt{OpenAI o1}. Costs reflect output tokens, excluding input prompts.}
        \label{fig:cost_analysis_boxplot}
    \end{subfigure}
    \caption{Comparison of time taken (a) and costs (b) using different LLMs and prompting strategies in \textsc{AIssistant}.}
    \label{fig:time_cost_comparison}
    \vspace{-0.4cm}
\end{figure*}

\subsection{Inference time and Computational Cost Analysis}

Figures~\ref{fig:time_duration_barchart} and~\ref{fig:cost_analysis_boxplot} show clear differences in inference time and cost between \texttt{OpenAI o1} and \texttt{GPT-4o-mini}. The maximum cost for \texttt{OpenAI o1} approaches \$0.90 USD, compared to only \$0.0037 USD for \texttt{GPT-4o-mini}, while differences in inference time are smaller. Review papers consistently required longer processing, with \texttt{GPT-4o-mini} sometimes taking more time for review workflows than \texttt{OpenAI o1} needed for perspective workflows. Inference times also increased in later \textsc{AIssistant} stages due to cumulative token growth across sequential agent outputs.

\section{Future Work \& Ethical Considerations}
\label{sec:future_work}

\noindent \textbf{Workflow Limitations.} While \textsc{AIssistant} supports perspective and review research and paper writing, several limitations remain. Iterative improvement across stages is limited, image generation is unsupported, and citation, formatting, and page-structure handling require stronger global agent control and more reliable inter-agent communication. Cross-domain LLM optimization is also unexplored, and using specialized LLMs for reasoning-intensive vs. routine tasks may improve subsection-level quality. The pipeline is sequential with fixed agents, whereas real research requires dynamic structure adaptation and iterative refinement. Even with a reasoning LLM like \texttt{OpenAI o1}, outputs may still include hallucinated citations, altered titles, or formatting errors, highlighting the need for human verification. Tables and structured formatting also require greater robustness. Our evaluation identifies trends but is insufficient for strong statistical claims; larger datasets and additional sub-domains are needed. Future work will expand both the dataset size and domain coverage for broader generalization.

\noindent\textbf{Ethical Considerations.} \textsc{AIssistant} reduces effort in scholarly workflow but may enable substandard or misleading outputs, potentially straining peer review. Human-AI collaboration can amplify biases from training data or prior outputs; however, we believe that the transparent LLM disclosure is essential. Misuse risks exist across domains, including chemistry, physics, cybersecurity, biomedicine, social sciences, and finance. Governance, bias mitigation, and accountability are critical to ensure outputs remain ethically sound and scientifically rigorous. Responsibility for ethical use is shared between system providers and humans, consistent with the EU AI Act~\cite{eu_ai_act}.


\section{Conclusion}

We presented \textsc{AIssistant}, an open-source Human--AI collaborative framework that accelerates data-science research workflows through guided, human-centered control rather than full automation. Using structured Human--AI collaboration, we evaluated usefulness, rigor, and workflow quality, finding both \texttt{GPT-4o-mini} and \texttt{OpenAI o1} effective, with \texttt{OpenAI o1} offering the strongest reasoning performance and favorable runtime–cost tradeoffs. Our HCI evaluation further demonstrated substantial efficiency gains in perspective and review workflows. Despite these benefits, gaps remain in clarity, soundness, and citation reliability, underscoring the need for structured human oversight. Within the broader AI for science landscape, \textsc{AIssistant} is the first modular pipeline explicitly designed for structured review and perspective generation in data science. Challenges persist in hallucinations, limited dynamic document adaptation, and constrained automated verification, highlighting opportunities for future improvement. Planned extensions include domain-specialized reasoning LLMs, more robust citation and formatting control, multimodal content support, and progress toward fully open-source scientific LLMs. Overall, \textsc{AIssistant} offers a scalable, accountable human--AI workflow that reduces repetitive effort while preserving research integrity and accelerating scientific discovery.

\section{Acknowledgments}

We thank the reviewers and the Program Chair for their time and feedback. We also acknowledge the TIB development team for the initial codebase and their technical guidance. This work was supported by the BMBF (KISSKI project, 01IS22093C) and the DFG (NFDI4DataScience consortium, project 460234259).


\bibliographystyle{IEEEtran}
\bibliography{references}

\end{document}